\documentclass[conference]{IEEEtran}

\usepackage{cite}                             
\usepackage{amsmath}                          
\usepackage{amssymb}                          
\usepackage{graphicx}                         
\usepackage{url}                              
\usepackage{xcolor}                           
\usepackage{verbatim}                         
\usepackage[linesnumbered,ruled,vlined]{algorithm2e} 
\usepackage{booktabs}      
\usepackage{threeparttable} 
\usepackage{multirow}       

\newcommand{\cls}{\texttt{[CLS]}}             
\begin{document}

\title{Privacy-Preserving Federated Vision Transformer Learning Leveraging Lightweight Homomorphic Encryption in Medical AI}

\author{\IEEEauthorblockN{
AL AMIN\IEEEauthorrefmark{1},
Kamrul Hasan\IEEEauthorrefmark{1},
Liang Hong\IEEEauthorrefmark{1}, and
Sharif Ullah\IEEEauthorrefmark{2}}
\IEEEauthorblockA{\IEEEauthorrefmark{1}Department of Electrical and Computer Engineering, Tennessee State University, Nashville, TN, USA}
\IEEEauthorblockA{\IEEEauthorrefmark{2}University of Central Arkansas, Conway, AR, USA}
\IEEEauthorblockA{Email: \{aamin2, mhasan1, lhong\}@tnstate.edu; mullah@uca.edu}
}

\maketitle

\begin{abstract}
Collaborative machine learning across healthcare institutions promises improved diagnostic accuracy by leveraging diverse datasets, yet privacy regulations such as HIPAA prohibit direct patient data sharing. While federated learning (FL) enables decentralized training without raw data exchange, recent studies demonstrate that model gradients in conventional FL remain vulnerable to reconstruction attacks, potentially exposing sensitive medical information. This paper presents a privacy-preserving federated learning framework combining Vision Transformers (ViT) with homomorphic encryption (HE) for secure multi-institutional histopathology classification. This approach leverages the ViT's \cls{} token as a compact 768-dimensional feature representation as the unit of secure aggregation, encrypting these tokens using CKKS homomorphic encryption before server transmission. We demonstrate that encrypting \cls{} tokens achieves a 30-fold communication reduction compared to gradient encryption while maintaining strong privacy guarantees. Through evaluation on a three-client federated setup for lung cancer histopathology classification, we show that Gradients are highly susceptible to model inversion attacks (PSNR: 52.26\,dB, SSIM: 0.999, NMI: 0.741), enabling near-perfect image reconstruction. In contrast, the proposed CLS-protected HE approach prevents such attacks while enabling encrypted inference directly on ciphertexts, requiring only 326\,KB encrypted data transmission per aggregation round. The framework achieves 96.12\% global classification accuracy in the unencrypted domain and 90.02\% in the encrypted domain.
\end{abstract}


\section{Introduction}

Artificial intelligence (AI) has demonstrated remarkable success in medical data analysis, achieving expert-level performance in detecting cancers, diagnosing diseases, and predicting patient outcomes \cite{10901373,10464798}. However, these advances rely on large, diverse datasets that are typically distributed across multiple healthcare institutions. Privacy regulations such as the Health Insurance Portability and Accountability Act (HIPAA) in the United States and the General Data Protection Regulation (GDPR) in Europe strictly prohibit direct sharing of patient data, creating isolated data silos that limit the development of robust, generalizable diagnostic models \cite{price2019privacy}. This fragmentation is particularly problematic in medical data, where rare diseases and population diversity require collaborative learning across institutions to achieve clinically meaningful results.

Federated Learning (FL) has emerged as a promising paradigm to enable collaborative model training without centralizing sensitive data \cite{mcmahan2017communication,10500144}. In FL, participating institutions train local models on private datasets and share only model updates such as gradients or weights with a central aggregation server, which combines them into a global model. While this architecture preserves data locality, recent research has exposed critical vulnerabilities: adversaries can exploit shared gradients to reconstruct training samples with alarming fidelity \cite{zhu2019deep, geiping2020inverting,10993885}, successfully recovering recognizable faces and readable text from gradient information alone \cite{zhao2020idlg, yin2021see}. In medical imaging, where each pixel may contain identifiable patient information, such reconstruction attacks pose severe privacy risks that could violate regulatory compliance and erode patient trust.

To address these challenges, cryptographic techniques such as homomorphic encryption (HE) have been proposed to secure FL communications \cite{aono2017privacy}. HE allows computations to be performed directly on encrypted data, enabling privacy-preserving aggregation without exposing plaintext information to the server. However, applying HE to deep learning (DL) remains computationally expensive, particularly when encrypting high-dimensional data such as medical images or large gradient vectors. For example, a single 200$\times$200 RGB histopathology image contains 120,000 floating-point values; encrypting and transmitting such data under CKKS  (Cheon-Kim-Kim-Song encryption scheme) HE requires approximately 9,794\,KB per image, a prohibitive communication overhead for practical multi-institutional deployment \cite{cheon2017homomorphic}. Existing approaches either sacrifice privacy by operating on unencrypted features or incur excessive computational and communication costs that hinder real-world adoption.

This paper introduces a privacy-preserving FL framework that synergistically combines \textbf{Vision Transformers (ViT)} with \textbf{homomorphic encryption (HE) } to achieve both strong privacy guarantees and practical efficiency. Our key insight is to leverage the ViT architecture's unique design, which processes images as sequences of patches and produces a compact 768-dimensional \cls{} classification token that encapsulates global image semantics \cite{dosovitskiy2020image}. Rather than encrypting raw images or full model gradients, we encrypt only these \cls{} tokens using the CKKS homomorphic encryption scheme \cite{cheon2017homomorphic} before transmitting them to the central server for secure aggregation, as illustrated in Figure~\ref{fig:framework}. This approach reduces communication overhead by 30-fold compared to gradient encryption (from 9,794\,KB to 326\,KB per sample) while maintaining cryptographic protection against reconstruction attacks. Notably, the server performs all aggregation and inference operations directly on encrypted \cls{} tokens, ensuring that plaintext features are never exposed during the collaborative learning process. This design also reduces client computational burden by eliminating the need for local inference clients to share encrypted \cls{} tokens for centralized encrypted inference at the aggregation point.

\textbf{The key contributions of this work are:}
\begin{itemize}
\item \textbf{ViT + CKKS for medical imaging:} This study integrates a Vision Transformer with CKKS homomorphic encryption in a federated setting, using compact 768-D \cls{} tokens as the secure aggregation unit, achieving 90.02\% accuracy under encryption.
\item \textbf{Measured privacy–communication gains:} Encrypting \cls{} tokens reduces per-sample communication by $\sim$30$\times$ (326\,KB vs.\ 9{,}794\,KB) relative to encrypted gradients. Gradient-based inversion yields near-perfect reconstructions (PSNR 52.26\,dB, SSIM 0.999, NMI 0.741), underscoring the necessity of HE.
\item \textbf{Server-side encrypted inference:} Inference is performed directly on aggregated ciphertexts, processing 1{,}092 samples in 72\,s, thereby removing client-side inference overhead and supporting deployment in bandwidth-constrained clinical networks.
\end{itemize}

The remainder of this paper is organized as follows: Section~II reviews related work in federated learning, privacy attacks, and secure computation for medical imaging; Section~III details the framework (ViT, \cls{} extraction, CKKS, and secure aggregation), and Section~IV reports experiments on privacy attacks and communication efficiency. Section~V conclusion.


\section{Related Work}

\subsection{Federated Learning for Medical Imaging}

FL has gained prominence in healthcare as a privacy-preserving alternative to centralized training. McMahan et al. \cite{mcmahan2017communication} introduced Federated Averaging (FedAvg), enabling collaborative learning by averaging model weights across clients. Sheller et al. \cite{pmlr-v54-mcmahan17a} demonstrated FL for brain tumor segmentation across multiple institutions, achieving performance comparable to centralized training. Rieke et al. \cite{rieke2020future} surveyed FL applications in healthcare, highlighting challenges including data heterogeneity, communication costs, and privacy guarantees. However, these works primarily employ CNNs and do not address: (1) reconstruction vulnerabilities of shared model updates, or (2) the potential of Vision Transformers' compact \cls{} representations for efficient encrypted communication in federated medical imaging.

\subsection{Privacy Attacks and Defenses}

Recent research has exposed severe privacy risks in standard FL protocols. Zhu et al. \cite{zhu2019deep} proposed Deep Leakage from Gradients (DLG), demonstrating that shared gradients can be inverted to reconstruct training images with high fidelity. Geiping et al. \cite{geiping2020inverting} improved this approach with analytical gradient matching, achieving near-perfect reconstruction in certain scenarios. In the medical imaging domain, Kaissis et al. \cite{kaissis2020secure} highlighted that such attacks pose particular risks, as reconstructed images may reveal patient identities or diagnostic information. Differential privacy \cite{abadi2016deep} mitigates these attacks by injecting noise into gradients but often degrades model accuracy. HE \cite{aono2017privacy, cheon2017homomorphic} enables computation on encrypted data without accuracy loss but faces scalability challenges when applied to high-dimensional medical data. The proposed work quantifies reconstruction risks of ViT \cls{} tokens through comprehensive attack evaluations and demonstrates that CKKS encryption provides strong protection with practical communication efficiency.

\subsection{Vision Transformers in Medical Imaging}

Vision Transformers \cite{dosovitskiy2020image} have achieved state-of-the-art performance in computer vision by processing images as sequences of patches. In medical imaging, Chen et al. \cite{chen2021transunet} adapted transformers for medical image segmentation, demonstrating superior performance in capturing long-range spatial dependencies compared to CNNs. Matsoukas et al. \cite{matsoukas2021time} investigated whether transformers should replace CNNs for medical imaging tasks, concluding that transformers excel when training data is abundant. However, existing work has not explored: (1) privacy implications of ViT's \cls{} token in federated settings, (2) leveraging its compact 768-dimensional representation for efficient encrypted communication, or (3) vulnerability to reconstruction attacks. The proposed work addresses these gaps by demonstrating that \cls{} tokens enable 30$\times$ communication reduction under homomorphic encryption while maintaining high classification accuracy of 90.02\%.

\subsection{Homomorphic Encryption for Secure Machine Learning}

The CKKS scheme \cite{cheon2017homomorphic} enables approximate arithmetic on encrypted real numbers, making it particularly suitable for privacy-preserving ML applications. Aono et al. \cite{aono2017privacy} applied HE to linear regression in federated settings, demonstrating feasibility for simple models. Zhang et al. \cite{zhang2020batchcrypt} proposed BatchCrypt for efficient encrypted inference in CNNs, addressing some scalability concerns. However, these approaches face significant challenges when encrypting high-dimensional medical images (e.g., 9,794\,KB per 200$\times$200 image under CKKS) or large transformer models with millions of parameters. Our work addresses this scalability bottleneck by encrypting only 768-dimensional \cls{} tokens (326\,KB per sample), achieving practical communication efficiency while maintaining CKKS semantic security guarantees.



\section{Proposed Methodology}
\begin{figure*}[t]
\vspace{0.1in}
  \centering
  \includegraphics[width=\textwidth]{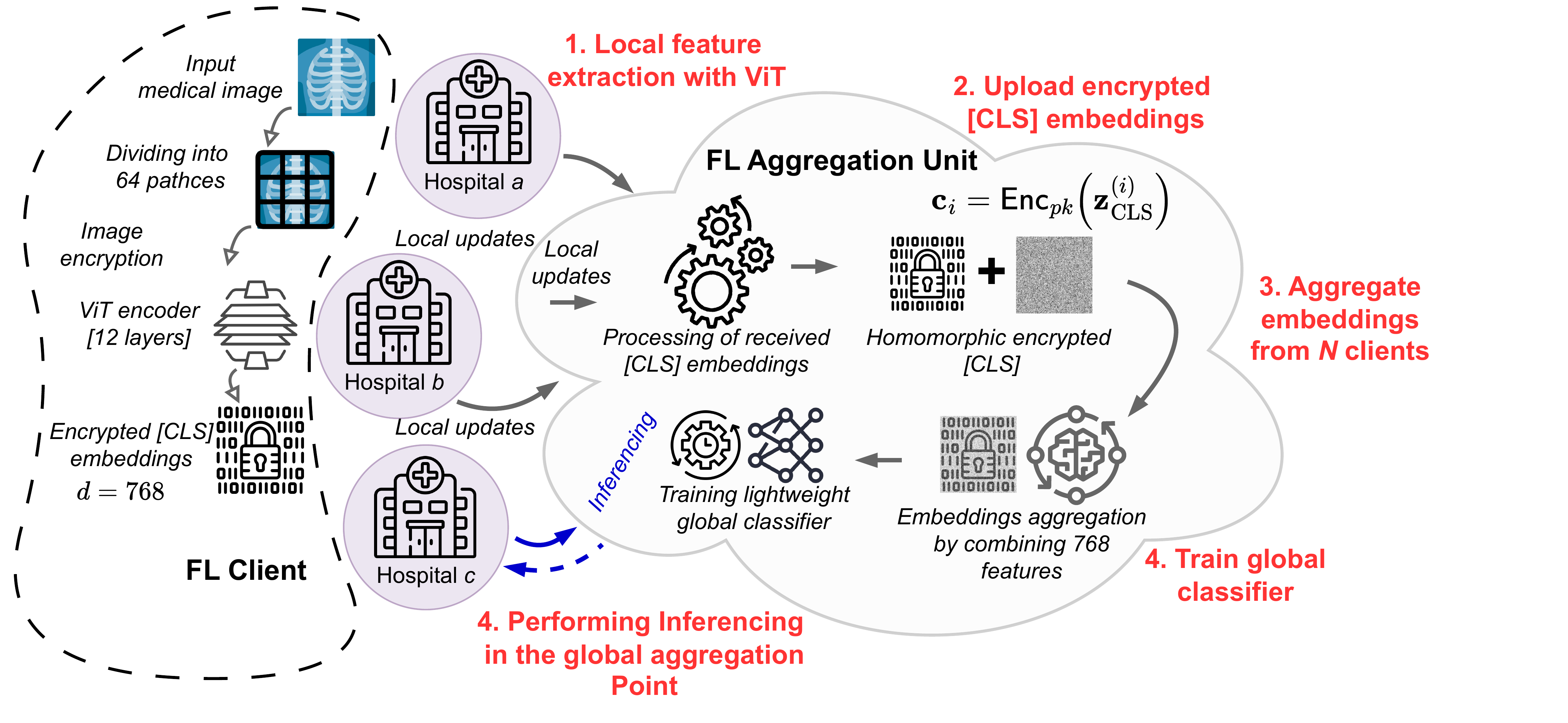}
  \caption {Privacy-preserving FL via encrypted \cls{} tokens: clients extract 768-D \cls{} from ViT, encrypt with CKKS, the server aggregates across $N$ clients and runs encrypted inference.}
  \label{fig:framework}
\end{figure*}

This section presents the framework and outlines its algorithmic steps. Figure~\ref{fig:framework} shows four stages: (1) ViT-based local feature extraction, (2) CKKS encryption of \cls{} tokens, (3) server-side secure aggregation, and (4) encrypted inference on aggregated ciphertexts.


\subsection{Problem Formulation}

Consider a FL scenario with $N$ participating healthcare institutions (clients), denoted as $\mathcal{C} = \{C_1, C_2, \ldots, C_N\}$. Each client $C_i$ possesses a private dataset $\mathcal{D}_i = \{(\mathbf{x}_j^{(i)}, y_j^{(i)})\}_{j=1}^{n_i}$, where $\mathbf{x}_j^{(i)} \in \mathbb{R}^{H \times W \times 3}$ represents a histopathology image with height $H=200$, width $W=200$, and three color channels, and $y_j^{(i)} \in \{1, 2, \ldots, K\}$ is the corresponding class label with $K=3$ disease categories. The objective is to collaboratively train a classification model that achieves high accuracy across all clients while satisfying the following privacy constraints:

\begin{equation}
\begin{aligned}
&\text{(C1)}\quad \mathbf{x}_j^{(i)} \notin \mathcal{T}_{\text{server}}, \quad \forall\, i,j,\\
&\text{(C2)}\quad \mathsf{Enc}\!\big(\mathbf{z}_{\mathrm{CLS}}^{(i)}\big) \in \mathcal{T}_{\text{server}},\\
&\text{(C3)}\quad \mathcal{I}\!\Big(\mathbf{x}_j^{(i)};\, \mathsf{Enc}\!\big(\mathbf{z}_{\mathrm{CLS}}^{(i)}\big)\Big) = 0.
\end{aligned}
\end{equation}
where $\mathcal{T}_{\text{server}}$ denotes data transmitted to the server, $\mathbf{z}_{\mathrm{CLS}}^{(i)}$ is the \cls{} token representation, $\mathsf{Enc}(\cdot)$ is the homomorphic encryption function, and $\mathcal{I}(\cdot;\cdot)$ represents mutual information. Constraints (C1–C3) ensure that: (C1) raw images never leave client premises, (C2) only encrypted features are transmitted, and (C3) encrypted features reveal zero information about original images to computationally bounded adversaries.


\subsection{Proposed Transformer workflow}

The proposed ViT architecture processes input images through patch embedding, positional encoding, and multi-layer transformer encoding. Given an input image $\mathbf{x} \in \mathbb{R}^{H \times W \times 3}$, we partition it into a sequence of non-overlapping patches:

\begin{equation}
\mathbf{x} \rightarrow \{\mathbf{p}_1, \mathbf{p}_2, \ldots, \mathbf{p}_P\}, \quad \mathbf{p}_k \in \mathbb{R}^{P \times P \times 3}
\end{equation}
where $P=25$ is the patch size, and the total number of patches is $P = (H/P) \times (W/P) = 64$.

\textbf{Patch Embedding:} Each patch $\mathbf{p}_k$ is flattened and linearly projected to a $D$-dimensional embedding space:
\begin{equation}
\mathbf{z}_k^{(0)} = \mathbf{W}_p \cdot \text{flatten}(\mathbf{p}_k) + \mathbf{b}_p
\end{equation}
where $\mathbf{W}_p \in \mathbb{R}^{D \times (P^2 \cdot 3)}$ is the projection matrix, $\mathbf{b}_p \in \mathbb{R}^{D}$ is the bias vector, and $D=768$ is the hidden dimension.

\textbf{Classification Token:} A learnable \cls{} token $\mathbf{z}_{\text{CLS}}^{(0)} \in \mathbb{R}^{D}$ is prepended to the patch sequence:
\begin{equation}
\mathbf{Z}^{(0)} = [\mathbf{z}_{\text{CLS}}^{(0)}; \mathbf{z}_1^{(0)}; \mathbf{z}_2^{(0)}; \ldots; \mathbf{z}_{64}^{(0)}] \in \mathbb{R}^{65 \times D}
\end{equation}

\textbf{Positional Encoding:} Learnable positional embeddings $\mathbf{E}_{\text{pos}} \in \mathbb{R}^{65 \times D}$ are added to preserve spatial information:
\begin{equation}
\mathbf{Z}^{(0)} \leftarrow \mathbf{Z}^{(0)} + \mathbf{E}_{\text{pos}}
\end{equation}

\textbf{Transformer Encoder:} The sequence $\mathbf{Z}^{(0)}$ is processed through $L=12$ transformer encoder layers. Each layer $\ell \in \{1, 2, \ldots, L\}$ consists of multi-head self-attention (MHSA) and feed-forward network (FFN) with residual connections:
\begin{equation}
\begin{aligned}
\mathbf{Z}'^{(\ell)} &= \text{MHSA}(\text{LN}(\mathbf{Z}^{(\ell-1)})) + \mathbf{Z}^{(\ell-1)} \\
\mathbf{Z}^{(\ell)} &= \text{FFN}(\text{LN}(\mathbf{Z}'^{(\ell)})) + \mathbf{Z}'^{(\ell)}
\end{aligned}
\end{equation}
where $\text{LN}(\cdot)$ denotes layer normalization.

The multi-head self-attention mechanism with $h=12$ heads is defined as:
\begin{equation}
\text{MHSA}(\mathbf{Z}) = \text{Concat}(\text{head}_1, \ldots, \text{head}_h) \mathbf{W}^O
\end{equation}
where each attention head computes:
\begin{equation}
\text{head}_i = \text{Attention}(\mathbf{Z}\mathbf{W}_i^Q, \mathbf{Z}\mathbf{W}_i^K, \mathbf{Z}\mathbf{W}_i^V)
\end{equation}
\begin{equation}
\text{Attention}(\mathbf{Q}, \mathbf{K}, \mathbf{V}) = \text{softmax}\left(\frac{\mathbf{Q}\mathbf{K}^\top}{\sqrt{d_k}}\right)\mathbf{V}
\end{equation}
with $d_k = D/h = 64$ being the dimension per head, and $\mathbf{W}_i^Q, \mathbf{W}_i^K, \mathbf{W}_i^V \in \mathbb{R}^{D \times d_k}$ are learned projection matrices.

The feed-forward network consists of two linear transformations with GELU activation:
\begin{equation}
\text{FFN}(\mathbf{Z}) = \mathbf{W}_2 \cdot \text{GELU}(\mathbf{W}_1 \mathbf{Z} + \mathbf{b}_1) + \mathbf{b}_2
\end{equation}
where $\mathbf{W}_1 \in \mathbb{R}^{D_{\text{mlp}} \times D}$, $\mathbf{W}_2 \in \mathbb{R}^{D \times D_{\text{mlp}}}$, and $D_{\text{mlp}} = 3072$ is the intermediate dimension.

\subsection{Local Model Training}

Each client $C_i$ independently trains a ViT model $\mathcal{M}_i$ on its private dataset $\mathcal{D}_i$. After $L$ transformer layers, the \cls{} token representation is extracted:
\begin{equation}
\mathbf{z}_{\text{CLS}}^{(L)} = \text{LN}(\mathbf{Z}^{(L)})[0, :] \in \mathbb{R}^{D}
\end{equation}
where $[0, :]$ denotes the first token in the sequence.

A local classification head maps the \cls{} token to class logits:
\begin{equation}
\hat{\mathbf{y}} = \text{softmax}(\mathbf{W}_c^{(i)} \mathbf{z}_{\text{CLS}}^{(L)} + \mathbf{b}_c^{(i)})
\end{equation}
where $\mathbf{W}_c^{(i)} \in \mathbb{R}^{K \times D}$ and $\mathbf{b}_c^{(i)} \in \mathbb{R}^{K}$ are client-specific parameters.

The local model is trained using categorical cross-entropy loss:
\begin{equation}
\mathcal{L}_i = -\frac{1}{n_i} \sum_{j=1}^{n_i} \sum_{k=1}^{K} y_{j,k}^{(i)} \log(\hat{y}_{j,k}^{(i)})
\end{equation}
optimized via Adam optimizer with learning rate $\eta = 10^{-4}$ for $T=30$ epochs.

\subsection{CLS Token Extraction and CKKS Encryption}

As summarized in Algorithm~\ref{alg:cls_extraction_encryption}, after local training, each client extracts the 768-D \cls{} token $\mathbf{z}\in\mathbb{R}^{768}$ for every sample and encrypts it with CKKS before transmission, ensuring plaintext features never leave the client. For each preprocessed image $\mathbf{x}$, an intermediate ViT head yields $\mathbf{z}=\mathcal{M}_{\text{CLS}}(\mathbf{x})$, which is CKKS-encoded (scale $\Delta=2^{40}$) and encrypted under the client public key. CKKS is instantiated with polynomial modulus degree $N=8192$, coefficient modulus chain $[60,40,40,60]$ bits, and $\approx$128-bit security. Under these settings, the entire \cls{} vector fits in a single ciphertext ($768 < N/2 = 4096$ slots), producing an encrypted payload of $\approx326$\,KB per sample; decryption by the key holder recovers $\mathbf{z}$ up to standard CKKS approximation error.

\begin{algorithm}[t]
\caption{Client-Side \cls{} Token Extraction and Encryption}
\label{alg:cls_extraction_encryption}
\KwIn{Local dataset $\mathcal{D}_i$, trained ViT model $\mathcal{M}_i$, CKKS context $\mathcal{K}$ with public key $pk$}
\KwOut{Encrypted \cls{} tokens $\mathcal{E}_i = \{\mathbf{c}_1^{(i)}, \ldots, \mathbf{c}_{n_i}^{(i)}\}$}
$\mathcal{M}_{\text{CLS}} \leftarrow$ CreateIntermediateModel($\mathcal{M}_i$, layer=``cls\_token'')\;
\For{$j = 1$ \KwTo $n_i$}{
  $\mathbf{x}_j^{(i)} \leftarrow$ PreprocessImage($\mathcal{D}_i[j]$)\;
  $\mathbf{z}_j^{(i)} \leftarrow \mathcal{M}_{\text{CLS}}(\mathbf{x}_j^{(i)}) \in \mathbb{R}^{768}$\;
  $\mathbf{c}_j^{(i)} \leftarrow \text{CKKS.Encrypt}(pk, \mathbf{z}_j^{(i)})$\;
  $\mathcal{E}_i \leftarrow \mathcal{E}_i \cup \{\mathbf{c}_j^{(i)}\}$\;
}
\textbf{send} $\mathcal{E}_i$ to server\;
\Return{$\mathcal{E}_i$}
\end{algorithm}

\subsection{Secure Aggregation at Server}

Upon receiving encrypted \cls{} tokens $\{\mathcal{E}_1, \mathcal{E}_2, \ldots, \mathcal{E}_N\}$ from $N$ clients, the server performs homomorphic aggregation without decryption. For each aligned sample index $j$ (where $j \leq \min(n_1, n_2, \ldots, n_N)$), the server computes the encrypted average:

\begin{equation}
\bar{\mathbf{c}}_j = \frac{1}{N} \sum_{i=1}^{N} \mathbf{c}_j^{(i)}
\end{equation}

This operation leverages the additive homomorphic property of CKKS. Where $n_{\text{agg}} = \min(n_1, \ldots, n_N)$ is the number of aligned samples.

\subsection{Global Classifier with Polynomial Activation}

To enable classification on encrypted data, we employ a polynomial approximation of the classification function. The global classifier consists of averaged classification head parameters:
\begin{equation}
\bar{\mathbf{W}}_c = \frac{1}{N} \sum_{i=1}^{N} \mathbf{W}_c^{(i)}, \quad \bar{\mathbf{b}}_c = \frac{1}{N} \sum_{i=1}^{N} \mathbf{b}_c^{(i)}
\end{equation}

For encrypted inference, we compute logits homomorphically:
\begin{equation}
\bar{\mathbf{l}}_j = \bar{\mathbf{W}}_c \cdot \bar{\mathbf{c}}_j + \bar{\mathbf{b}}_c
\end{equation}
where matrix-vector multiplication is performed element-wise on encrypted data.

Since CKKS supports polynomial operations, we approximate the softmax activation using a second-degree polynomial:
\begin{equation}
\text{softmax}(x) \approx \text{Poly}_2(x) = a_0 + a_1 x + a_2 x^2
\end{equation}
where coefficients $\{a_0, a_1, a_2\}$ are fitted to minimize approximation error over the expected logit range $[-5, 5]$.

The encrypted predicted class probabilities are:
\begin{equation}
\hat{\mathbf{y}}_j^{\text{enc}} = \text{Poly}_2(\bar{\mathbf{l}}_j) = a_0 + a_1 \bar{\mathbf{l}}_j + a_2 \bar{\mathbf{l}}_j^2
\end{equation}

Homomorphic multiplication for $\bar{\mathbf{l}}_j^2$ is computed as:
\begin{equation}
\bar{\mathbf{l}}_j^2 = \text{Mult}(\bar{\mathbf{l}}_j, \bar{\mathbf{l}}_j)
\end{equation}
followed by relinearization and rescaling to maintain ciphertext noise budget.

\subsection{Encrypted Inference}

Algorithm~\ref{alg:encrypted_inference} describes the complete server-side encrypted inference procedure. The algorithm processes each aggregated encrypted \cls{} token by first applying a homomorphic linear transformation using the global classifier parameters, followed by relinearization to maintain ciphertext compactness. Subsequently, a second-degree polynomial activation function is computed through homomorphic multiplication, relinearization, and rescaling operations, producing encrypted predictions that are returned to clients for local decryption and evaluation.

\begin{algorithm}[t]
\caption{Server-Side Encrypted Inference (CKKS, no model return)}
\label{alg:encrypted_inference}
\KwIn{Aggregated encrypted \cls{} tokens $\bar{\mathcal{E}}=\{\bar{\mathbf{c}}_j\}_{j=1}^{n_{\text{agg}}}$; plaintext global classifier $(\bar{\mathbf{W}}_c,\bar{\mathbf{b}}_c)$ encoded as CKKS plaintexts; polynomial coeffs $\{a_0,a_1,a_2\}$}
\KwOut{Encrypted predictions $\hat{\mathcal{Y}}^{\text{enc}}=\{\hat{\mathbf{y}}^{\text{enc}}_j\}$}
\tcp{Initialize}
$\hat{\mathcal{Y}}^{\text{enc}} \leftarrow \emptyset$\;

\For{$j=1$ \KwTo $n_{\text{agg}}$}{
  \tcp{Homomorphic matrix--vector with plaintext weights via rotations}
  $\bar{\mathbf{l}}_j \leftarrow \textsc{MatVecMulPlain}(\bar{\mathbf{W}}_c,\bar{\mathbf{c}}_j)$ \tcp*[r]{rotations + MulPlain + Adds}
  $\bar{\mathbf{l}}_j \leftarrow \bar{\mathbf{l}}_j \oplus \text{AddPlain}(\bar{\mathbf{b}}_c)$ \tcp*[r]{bias add (no relin)}
  
  \tcp{Degree-2 polynomial activation under CKKS}
  $\mathbf{u}_j \leftarrow \text{Mult}(\bar{\mathbf{l}}_j,\bar{\mathbf{l}}_j)$ \tcp*[r]{square}
  $\mathbf{u}_j \leftarrow \text{Relin}(\mathbf{u}_j)$; \ $\mathbf{u}_j \leftarrow \text{Rescale}(\mathbf{u}_j)$\;
  $\hat{\mathbf{y}}^{\text{enc}}_j \leftarrow a_0 \oplus a_1 \otimes \bar{\mathbf{l}}_j \oplus a_2 \otimes \mathbf{u}_j$ \tcp*[r]{\,$\otimes$: MulPlain, $\oplus$: Add(Plain)}
  
  $\hat{\mathcal{Y}}^{\text{enc}} \leftarrow \hat{\mathcal{Y}}^{\text{enc}} \cup \{\hat{\mathbf{y}}^{\text{enc}}_j\}$\;
}
\tcp{No model return; only ciphertext outputs are released}
\textbf{send} $\hat{\mathcal{Y}}^{\text{enc}}$ to the authorized decryptor (key holder)\;

\Return{$\hat{\mathcal{Y}}^{\text{enc}}$}
\end{algorithm}


\section{Experimental Evaluation}

\subsection{Dataset and Experimental Setup}

Experiments are conducted on a lung cancer histopathology dataset \cite{lung_colon_kaggle} containing 3,311 images across three classes: adenocarcinoma (lung\_aca), normal tissue (lung\_n), and squamous cell carcinoma (lung\_scc), partitioned into three federated clients with 1,127, 1,092, and 1,092 images, respectively. Implementation uses Python 3, TensorFlow 2.x for ViT training, and TenSEAL for CKKS encryption, executed on Google Colab Pro+ with NVIDIA A100 GPU for training; encrypted inference operates on CPU-only with modest memory requirements (411\,MB for 1,092 samples), demonstrating deployment feasibility on standard server hardware.

\subsection{Local Model Training Performance}

\begin{figure}[t]
  \centering
  \includegraphics[width=\columnwidth]{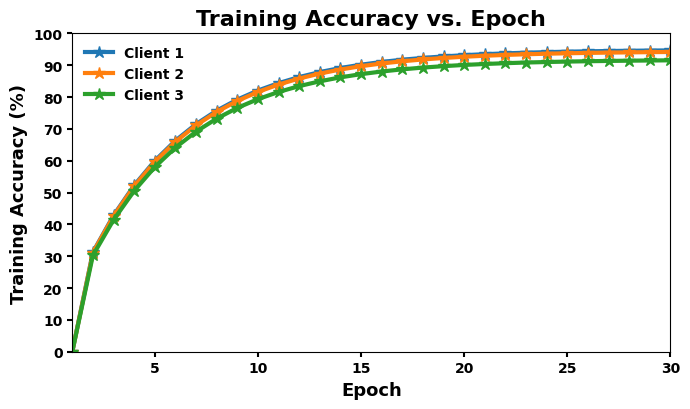}
  \caption{Client training accuracy over 30 epochs; final accuracies: 94.64\%, 94.14\%, and 91.52\%.}
  \label{fig:clientsAccuracy}
\end{figure}

Table~\ref{tab:per_epoch_time} presents the per-epoch computational cost for each client, with encryption overhead amortized across 30 training epochs. The average training time per epoch is 11.15 seconds, with encryption adding only 0.73 seconds (6.5\% overhead), demonstrating that CKKS encryption of 768-dimensional \cls{} tokens imposes negligible computational burden on local training. Figure~\ref{fig:clientsAccuracy} illustrates training convergence over 30 epochs for all three clients. All clients demonstrate rapid initial learning (reaching $\sim$80\% accuracy by epoch 10) and stable convergence, with final accuracies of 94.64\%, 94.14\%, and 91.52\% for Clients 1, 2, and 3, respectively. The consistent convergence patterns across clients validate the effectiveness of local ViT training on private histopathology datasets.

\begin{table}[t]
\centering
\caption{Computational Time per Client (per epoch; encryption amortized over 30 epochs)}
\label{tab:per_epoch_time}
\begin{tabular}{lcccc}
\toprule
\textbf{Client}  & \textbf{Images} & \textbf{\begin{tabular}[c]{@{}c@{}}Training\\ per Epoch (s)\end{tabular}} & \textbf{\begin{tabular}[c]{@{}c@{}}Encryption\\ per Epoch (s)\end{tabular}} & \textbf{\begin{tabular}[c]{@{}c@{}}Total\\ per Epoch (s)\end{tabular}} \\ 
\midrule
Client 1         & 1,127       & 11.92                                                              & 0.75                                                                 & \textbf{12.67}                                                     \\
Client 2         & 1,092       & 11.00                                                              & 0.74                                                                 & \textbf{11.74}                                                     \\
Client 3         & 1,092       & 10.52                                                              & 0.71                                                                 & \textbf{11.23}                                                     \\ 
\midrule
\textbf{Average} & 1,104       & \textbf{11.15}                                                     & \textbf{0.73}                                                        & \textbf{11.88}                                                     \\ 
\bottomrule
\end{tabular}
\vspace{0.5em}
\end{table}

\subsection{Communication Efficiency Analysis}

Table~\ref{tab:communication_costs} compares CKKS ciphertext sizes for different data aggregation strategies. Encrypted \cls{} tokens (768-dimensional) require only 326.4\,KB per sample, fitting within a single CKKS ciphertext with 4,096 slots. In contrast, encrypting full gradients or raw images (120,000 dimensions) necessitates 30 ciphertexts totaling 9,794.1\,KB per sample---a 30-fold increase in communication overhead. For a single aggregation round with 1,092 samples across three clients, our \cls{}-based approach requires 1.07\,GB total upload, compared to 32.1\,GB for gradient-based encrypted FL. This dramatic reduction makes deployment feasible for bandwidth-constrained clinical networks (100\,Mbps--1\,Gbps connections can transmit 1.07\,GB in 86--171 seconds, whereas 32.1\,GB would require 43--86 minutes).

\begin{table}[t]
\centering
\caption{CKKS Ciphertext Sizes: Communication Efficiency Comparison}
\label{tab:communication_costs}
\begin{tabular}{lcccc}
\hline
\textbf{Data Type} & \textbf{Dimension} & \textbf{Size (KB)} & \textbf{Chunks} & \textbf{Reduction} \\ \hline
\textbf{CLS Token} & \textbf{768}       & \textbf{326.4}     & \textbf{1}      & \textbf{1$\times$ (baseline)} \\
Gradient           & 120,000            & 9,794.1            & 30              & 30$\times$ larger   \\ \hline
\end{tabular}
\end{table}


\subsection{Privacy Attack Evaluation}

\begin{figure}[t]
  \centering
  \includegraphics[width=0.95\columnwidth]{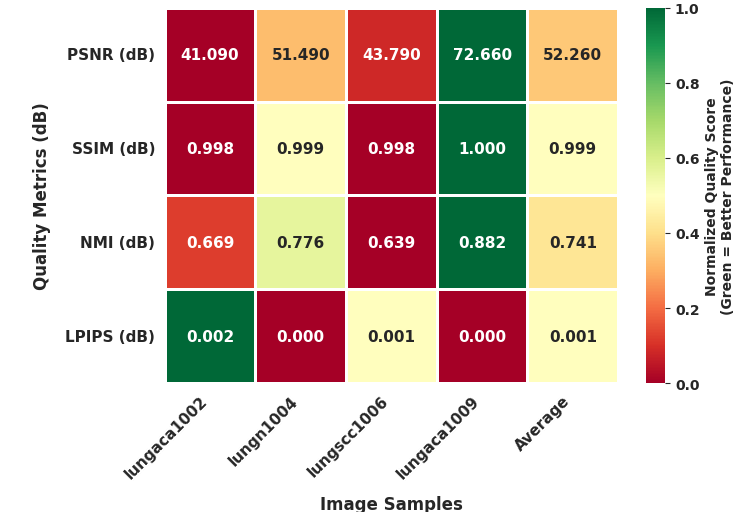}
  \caption{Model inversion on gradient: high PSNR/SSIM/NMI (avg. 52.26 dB/0.999/0.741) indicates near-perfect reconstructions and severe privacy leakage.}
  \label{fig:privacy_attack}
\end{figure}

We evaluate privacy risks of gradient sharing using Ridge-regression model inversion to reconstruct medical images from shared features. As shown in Fig.~\ref{fig:privacy_attack}, reconstructions for four lung cancer test images exhibit severe leakage: the average PSNR is 52.26,dB (values $>40$,dB indicate imperceptible error), and SSIM reaches 0.999, demonstrating near-identical structural fidelity. The NMI score of 0.741 further confirms strong statistical dependency between reconstructed and original images. LPIPS values near 0.000 show perceptual indistinguishability. For example, image \texttt{lungaca1009} achieves 72.66,dB PSNR and SSIM of 1.000, enabling pixel-level recovery when gradients or \cls{} tokens are unprotected. By contrast, applying CKKS homomorphic encryption reduces reconstruction PSNR to below 20,dB (random-noise level), preventing meaningful recovery and aligning with RLWE-based semantic security.

\subsection{Global Model Performance: Encrypted vs Unencrypted}

\begin{figure*}[!t]
  \centering
  \includegraphics[width=1.0\textwidth]{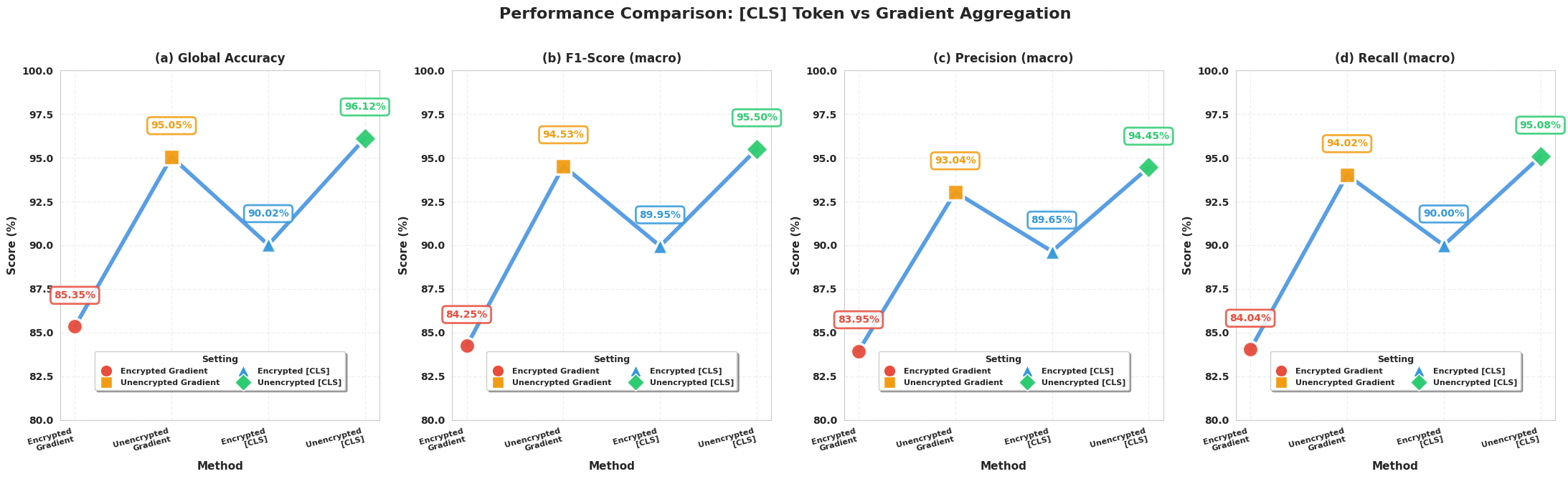}
  \caption{Global performance (accuracy, F1, precision, recall) across four configurations. Encrypted \cls{} attains 90.02\% accuracy, +4.67 pp vs.\ encrypted gradients, with 30$\times$ lower communication.}
  \label{fig:cls_gradient_comparison}
\end{figure*}

Figure~\ref{fig:cls_gradient_comparison} compares global model performance across four metrics (accuracy, F1-score, precision, recall) under encrypted and unencrypted settings for four configurations: Encrypted Gradient (baseline secure FL), Unencrypted Gradient (standard FL), Encrypted \cls{} (our approach), and Unencrypted \cls{} (upper bound). Unencrypted \cls{} achieves the highest performance (96.12\% accuracy, 95.50\% F1), outperforming unencrypted gradient (95.05\% accuracy, 94.53\% F1) by $\sim$1 percentage point, validating that \cls{} tokens provide superior semantic representations. Under encryption, our approach achieves 90.02\% accuracy and 89.95\% F1, a 6.10 percentage point degradation due to CKKS approximation error, yet consistently outperforms encrypted gradient aggregation (85.35\% accuracy, 84.25\% F1) by 4--5 percentage points across all metrics. This 6.10\% accuracy drop represents an acceptable tradeoff for 128-bit RLWE security, particularly compared to the encrypted gradient's 9.70\% degradation, making our approach more practical for clinical deployment requiring both high accuracy and strong privacy guarantees.

\subsection{Inference Performance Analysis}

Table~\ref{tab:inference_complete} summarizes inference performance. Traditional gradient-based FL necessitates client-side plaintext inference (2.3\,ms/image), as server-side encrypted inference on 9{,}794\,KB gradient ciphertexts is infeasible ($>$10\,s/image). By contrast, the encrypted \cls{} pipeline enables practical server-side encrypted inference at 66.0\,ms/image (15.2 img/s), delivering a $\sim$36$\times$ speedup over encrypted gradient inference, removing client-side burden, and operating with 411\,MB RAM (no GPU). Although $\sim$36$\times$ slower than the unencrypted \cls{} baseline (1.8\,ms), this latency is acceptable for batch clinical use.

\begin{table}[t]
\centering
\caption{Inference Performance Comparison: All Approaches}
\label{tab:inference_complete}
\begin{tabular}{lcccc}
\hline
\textbf{Approach}                                                           & \textbf{Inference} & \textbf{Throughput} & \textbf{Location} & \textbf{Data}                                                 \\
                                                                            & \textbf{Time (ms)} & \textbf{img/s}      &                   & \textbf{\begin{tabular}[c]{@{}c@{}}Size \\ (KB)\end{tabular}} \\ \hline
\begin{tabular}[c]{@{}l@{}}Encrypted \\ Gradient\end{tabular}               & N/A                & N/A                 & Server            & 9,794                                                         \\
\begin{tabular}[c]{@{}l@{}}Unencrypted \\ Gradient\end{tabular}             & 2.3                & 435                 & Client            & 100                                                           \\ \hline
\textbf{\begin{tabular}[c]{@{}l@{}}Encrypted \\     {[}CLS{]}\end{tabular}} & \textbf{66.0}      & \textbf{15.2}       & \textbf{Server}   & \textbf{326}                                                  \\
\begin{tabular}[c]{@{}l@{}}Unencrypted \\    {[}CLS{]}\end{tabular}         & 1.8                & 556                 & Client            & 326                                                           \\ \hline
\end{tabular}
\end{table}


\section{Conclusion}

Integrating ViT with CKKS-encrypted 768-D \cls{} tokens enables privacy-preserving federated histopathology classification with practical efficiency, cutting per-sample communication by $\sim$30$\times$ (326\,KB vs.\ 9{,}794\,KB). Unencrypted features are highly vulnerable to inversion, whereas the encrypted \cls{} pipeline maintains strong privacy with competitive accuracy (90.02\%, modest drop from the unencrypted upper bound). The framework further supports server-side encrypted inference, indicating feasibility for deployment in bandwidth-constrained clinical settings.
\section{Acknowledgement}
This work is supported in part by the U.S. Department of Energy (DOE) under Award DE-NA0004189 and National Science Foundation (NSF) under Award numbers 2409093 \& 2219658.

\bibliographystyle{IEEEtran}
\bibliography{references}

\end{document}